\pdfoutput=1
\documentclass[sigconf,nonacm]{acmart}
\usepackage{graphicx} 
\graphicspath{{figures/}}
\usepackage{booktabs} 
\usepackage{array} 
\usepackage{caption} 
\usepackage[title]{appendix}
\usepackage{algorithm}
\usepackage{algpseudocode}
\usepackage{amsmath}
\usepackage{amsfonts}
\usepackage{multirow}
\usepackage{tablefootnote}
\usepackage{wrapfig}
\usepackage{arydshln}
\usepackage{url}
\usepackage{subcaption}
\usepackage{float}
\begin{document}
\fancyhead{}

\title{CognitionNet: A Collaborative Neural Network for Play Style Discovery in Online Skill Gaming Platform}

\author{Rukma Talwadker}
\email{rukma.talwadker@games24x7.com}
\affiliation{%
  \institution{Artificial Intelligence and Data Science, Games24x7}
  \country{India}
  }
\author{Surajit Chakrabarty}
\email{surajit.chakrabarty@games24x7.com}
\affiliation{%
  \institution{Artificial Intelligence and Data Science, Games24x7}
  \country{India}
  }
\author{Aditya Pareek}
\email{aditya.pareek@gaes24x7.com}
\affiliation{%
  \institution{Artificial Intelligence and Data Science, Games24x7}
  \country{India}
  }   
\author{Tridib Mukherjee}
\email{tridib.mukherjee@games24x7.com}
\affiliation{%
  \institution{Artificial Intelligence and Data Science, Games24x7}
  \country{India}
  }
\author{Deepak Saini}
\email{deepak.saini@games24x7.com}
\affiliation{%
  \institution{Product Delight, Games24x7}
  \country{India}
  }

\begin{abstract}
Games are one of the safest source of realizing self-esteem and relaxation at the same time~\cite{sg5,games_vs_others}. An online gaming platform typically has massive data coming in, e.g., in-game actions, player moves, clickstreams, transactions etc. It is rather interesting, as something as simple as data on gaming moves can help create a psychological imprint of the user at that moment, based on her impulsive reactions and response to a situation in the game. Mining this knowledge can: (a) immediately help better explain observed and predicted player behavior; and (b) consequently propel deeper understanding towards players' experience, growth and protection. 

To this effect, we focus on discovery of the ``\textbf{game behaviours}'' as micro-patterns formed by continuous sequence of games and the persistent ``\textbf{play styles}'' of the players' as a \textit{sequence of such sequences} on an online skill gaming platform for Rummy. The \textit{complex} sequences of \textit{intricate} sequences is analysed through a novel \textit{collaborative} two stage deep neural network, \textit{CognitionNet}. The first stage focuses on mining game behaviours as cluster representations in a latent space while the second \textit{aggregates} over these micro patterns (e.g., transitions across patterns) to discover play styles via a supervised classification objective around player engagement. The dual objective allows CognitionNet to reveal several player psychology inspired decision making and  tactics. 
To our knowledge, this is the first and  \textit{one-of-its-kind} research to fully automate the discovery of: (i) player psychology and game tactics from telemetry data; and (ii) relevant diagnostic explanations to players' engagement predictions. The collaborative training of the two networks with differential input dimensions is enabled using a novel formulation of ``bridge loss''. The network plays pivotal role in obtaining homogeneous and consistent play style definitions and significantly outperforms the SOTA baselines wherever applicable.
\end{abstract}

\maketitle

\section{Introduction}
Social interaction and community affairs serve as the backdrop for satiating the esteem needs by exerting and validating prestige, personal worth, social acceptance, respect \& recognition requirements of an individual. As per the research \cite{sg5,games_vs_others}, games are one of the safest source of realizing self-esteem and relaxation at the same time. Online skill gaming platforms also have a responsibility to facilitate a competitive playing experience while preventing the situations of player over-indulgence. Games involve several crucial decision tactics involving what type of game to choose, what kind of move to make, whether to drop from a game on evaluating a \textit{no-win} or a \textit{less probable win} situation etc.. Evaluation of  players' choices against the various game situations can yield insights on player's play styles. This discovery can facilitate towards delivering a purposeful learning and growth experience to its players while ensuring player protection. In this paper we focus on discovering ``\textit{game behaviours}'' of players on the basis of their decisions, on an online skill gaming platform for \textit{Rummy}, as an example. Further to this, we challenge ourselves towards the discovery of their inherent game ``\textit{play styles}'' which is a compilation of their game behaviours over a sustained period of time. \footnote{This is a PRE-PRINT version of the work accepted for publication at ACM KDD'22. The final version is published by ACM and is available at: https://doi.org/10.1145/3534678.3539179}

This problem poses a number of significant \textit{challenges}. \textbf{Firstly} it's a problem of 3-dimensional data analysis in handling sequences of sequences: (i) game specific dynamics; (ii) continuous streak of games; and (iii) the longitudinal sequence of streaks over time. 
\textbf{Second}, there is an explosion of search space due to all possible combinations across these dimensions (e.g., multiple different game choices in a game streak). and there maybe overlapping of psychological imprints depending on the journey and experience over time. 
\textbf{Finally}, there is orthogonality between game play style and game outcome, e.g., a player in a losing streak can show assured and confident game play style while somebody winning may actually show unnecessary aggression, and vice versa.
Hence, there is no such dimension to pivot to discover game play styles. Prior work has typically focused on either---(a) quantitative analysis in these dimensions individually (e.g., game specific dynamics~\cite{pakddgames,kddgames}, temporal evolution~\cite{cods_games}) without any grounding on the psychological imprint; or conversely, (b) empirical psychological studies~\cite{caillois,MALONE1981333,big5,bighex} done on limited sets of players, without implicitly basing on any telemetry data. We contend to fill this key gap in this paper.

Intuitively, the psychological imprints can be formed primarily from the first two dimensions while the macro play styles stem from the aggregation of these imprints in the third dimension. usually, these have an impact on the next (future) observable behavior on the platform (e.g., engagement). 
 We propose a novel framework, ``CognitionNet'', for discovering these game play styles while having a dual objective of predicting observable behavior in terms of engagement. At a high level, CognitionNet is an assembly of two neural networks, where, the first network is focused on interpreting the first two dimensions into ``micro patterns''.  The second network has to leverage the \textit{collective wisdom} of these micro patterns (i.e., third dimension) for a player to mine her play style, while grounding that towards predicting the engagement. These networks collaborate to refine the representations of the micro patterns and create robust definitions of play styles. However, since the input dimensions differ, we propose a novel formulation for loss back-propagation, referred to as the \textit{Bridge loss} for the necessary collaboration. This eliminates dependency on a fixed network architecture for the second network, thereby enabling domain specific customization (e.g., different observable behavior prediction than engagement). It opens up a new possibility of collaborative training of two dissimilar and different dimensional input based networks. 
\footnote{We make the network and our training data publicly available \cite{github} for reproducibility and further research on discovery of actionable explainations on sequence of sequences in the related domains.}

  \begin{table*}
  \centering
  \footnotesize
 \begin{tabular}{||c| c| c|c| c|c| c||}
 \hline
  Model &  Sustainer R mean\%  & Sustainer P mean\% &  Burnout R mean\%  & Burnout P mean\% &  Churnout R mean\%  & Churnout P mean\%\\
 \hline  
  Random Forest(RF)& 18.2 & 39.2 & 10.4  & 57.3 & 89.1 & 60.01 \\
  XGBoost & 5.9 & 33.3 & 3.2 & 50.4 & 96.1 & 58.4\\
  SVM & 23.1 & 10.9 & 10.8 & 37.3 & 84.5 & 61.2  \\
  ANN & 17.3 & 38.9 & 5.6 & 40.7 & 89.7 & 59.2  \\
  \hline
\end{tabular}
 \caption{Performance of Various Classifiers on Engagement Class Prediction (mean - 5 runs): P - Precision, R - Recall} 
 \vspace{-0.2in}
 \label{tab:background_PR}
\end{table*}
 
\subsection{Background and Motivation:}
\label{background}
 The business observed that amongst the highly engaged, high skilled and regular players of the game, only a fraction of the players continued to engage with the same intensity month over month. There was curiosity in understanding why this is so and how we can improve the overall experience of these players. The category of players who continued to engage in a sustained manner are referred to as the \textbf{Sustainers}. Another fraction of players showed symptoms of poor engagement over a period of time and are termed as the \textbf{Burnouts}. The burnouts did continue playing, however, their stakes in the cash games substantially reduced along with less frequent visits to the game platform. The third category of players were termed as the \textbf{Churnouts}, which post heavy engagement for a sustained interval of time, completely dropped off from the platform, not to return back. 
 
\begin{wrapfigure}{l}{0.5\linewidth}
  \includegraphics[width=\linewidth]{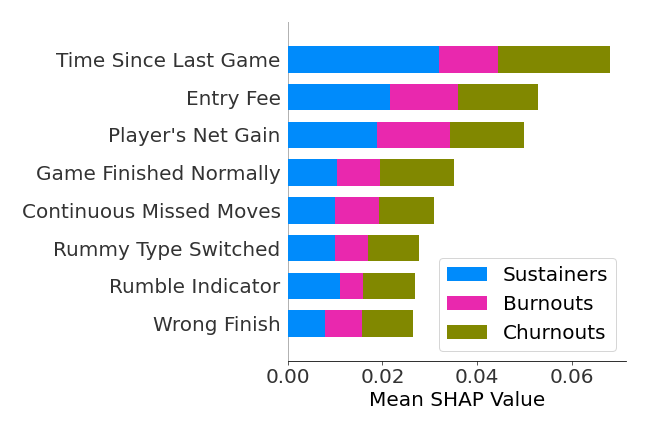}
  \caption{Random Forest Explainations using SHaP Model}
  \label{fig:SHAP}
\end{wrapfigure}
 We had access to the 3-dimensional game play data of these players for the entire period of the year 2021 along with their labels (Sustainer, Burnout and Churnout). We obtained  sequences for about 3 months of their respective game play periods until just before the event of burnout or churnout was observed by the business. We had balanced number of samples across the three classes. We created several aggregation based derived features on the various game play features, captured both at the level of a particular sequence and over all non-continuous sequences. Table \ref{tab:background_PR} shows precision recall of the various \textit{off the shelf} supervised classifiers including a simple 3 hidden layer based MLP neural network (ANN). We observed that maximum recall was only around 18\% and 10\% for the sustainers, burnouts respectively. This alluded to the fact that either the feature space is inadequate or the aggregations are probably not the right way to proceed.
 

Figure \ref{fig:SHAP}  highlights the feature importance over the random forest based model using SHaP  (SHapley Additive exPlanations) \cite{SHAP}. The most influential variables for prediction were her patience (Time Since Last Game), willingness to bet (Entry Fee) and payoff the player receives (Player's Net Gain). The model finds it hard to raise accuracy beyond ~60\% (which mostly comes from churnouts) as the key variables it looks at are more related to the monetary attitude of the player, instead of the gameplay specific variables like judgement on dropping a bad hand (drop adherence), etc. Since the features are at an aggregate level, model is not able to comprehend further on the progressive nature of their choices which could be an important differentiator. 
This motivated us to build a robust and differentiated neural network based architecture which could leverage players' game play data to discover existence of play styles via the pivot of their engagement classes.

\vspace{-0.1in}
\subsection{CognitionNet Contributions:} 
 CognitionNet breaks the myth that deep learning frameworks compromise human interpretable and actionable explainability over prediction accuracy.
 In doing so, CognitionNet makes the following \textbf{contributions}:

\noindent \textbf{1.} To the best of our knowledge \textit{CognitionNet} is the first network which \textit{collaboratively} trains the two networks of different dimensional inputs to model sequence of sequences. 
\\
\noindent \textbf{2.} The clustered micro-patterns mined by the first network reveal interesting \textit{Game Behaviours}, while the second network stitches these micro patterns to discover Players' \textit{Play Styles}. In doing so we combine game specific domain knowledge with research findings in the domain of player psychology.
\\
\noindent \textbf{3.} CognitionNet introduces a novel formulation of \textit{bridge loss} thereby eliminating dependency on fixed network architecture for the second network, thereby enabling domain specific customization. 

 The rest of the paper is organized as follows: We first review the existing literature in Section \ref{sec:related}. We introduce CognitionNet and the bridge loss formulation in the Section \ref{sec:cognitionNet} followed by a detailed discussion of the clasifier predictions and the discovered  play styles.

\vspace{-0.1in}
\section{Related Work}
\label{sec:related}
\noindent\textbf{Time Series Clustering Techniques:} For mining micro patterns from individual sequences across players we revisited the prior art in time series based clustering representations. We primarily looked at unsupervised methods of obtaining stable cluster representations. Raw-data-based methods mainly modify the distance function between the two time series. k-DBA algorithm  \cite{petitjean}, K-Spectral Centroid (K-SC) method \cite{yang}, k-shape that considers the shapes of the time series \cite{k-shape} were highly cited. However, they tend to be quite domain specific. Feature-based methods extract feature representations of input time series, to mitigate the impact of noise. \cite{guo} used independent component analysis, \cite{u-shapelet} proposed u-shapelet to learn local patterns. \cite{icjai1,icdm2} iteratively adopted local learning. These techniques mostly adopted linear features, while real time series tend to be non-linear \cite{10,11,12}. 
Feature extraction being merely a a pre-processing step, it may not be a channelized way of getting appropriate clusters and may appear as a disjoint step. The Deep Temporal Clustering Representation (\textbf{DTCR}) technique \cite{DTCR} integrates the temporal reconstruction and K-means objective into the seq2seq model. This approach claims improved cluster structures and cluster-specific temporal representations. Some of the  prior approaches leading up to this direction are \cite{dtc,spectral}. In CognitionNet, we build on top of the work presented in \cite{DTCR}. 

\noindent\textbf{Explainability in AI (XAI):} \cite{xai_study} presents a detailed survey of the various XAI techniques. Post-hoc methods approximate the behavior of a model by extracting relationships between feature values and predictions and has been mostly adopted in the area of computer vision. \cite{iccv} talks about explainability on images using a classifier. In the case of RNN based networks, attention mechanisms are being primarily suggested because they are embedded in the structure of recurrent networks and the explainability they offer is available directly at the end of the learning phase \cite{xai44}. \cite{xai-kim} proposes the use of SHapley Additive exPlanations (SHAP) algorithm \cite{SHAP}. \cite{LIME} proposes perturbation on the data to decode model's feature weight matrix. In CognitionNet we focus on behaviour discovery which can be actioned on. We leverage existing work wherever applicable.

\noindent\textbf{Cognitive Sciences and Player Psychology:} There has been a vast research, both in terms of understanding the psychology of players who exhibit gambling traits \cite{braverman,shaw} and also ways to detect them via ML\cite{percy,suriadi, gecco, akhter}. Some aspects of these questions have been addressed in player research \cite{caillois,MALONE1981333,big5,bighex}. We still lack comprehensive answers that adequately account for the role of a player's \textit{ludic habitus} – their past experiences, knowledge, and behavioural traits in manifestation of their individual play styles. Skill based cash games are at an unique intersection of understanding user's psychology from their telemetry data. These platforms contain: player's game play and in-game decision making information, game outcome as a win or a loss, and the monetary impact. 

\noindent\textbf{Collaborative Networks:} Collaborative learning has been exploited in many artificial intelligence areas such as language recognition, computer vision and expert systems including recommender systems. In \cite{google_collab} multiple classifier heads of the same network are simultaneously trained on the same training data to improve
generalization for image classification.  Different types of collaborative training including auxiliary training \cite{conv_deep}, multi-task learning \cite{multi_task},
and knowledge distillation \cite{distil} have been proposed in the literature. In the field of recommendation systems collaborative filtering with matrix factorization and augmentation of the learning feature space with latent features obtained from varied user-item interactions have been proposed \cite{user_collab}. None of these works are directly applicable to us, as these work on identical data dimensions throughtout.

\noindent\textbf{3-Dimensional Data Modeling in Language Models:} An evidence of 3-dimensional data modeling is seen in the domain of language models. Each game in our context can be associated with a word in a document. Sequence of games can be associated to a sentence.  Building on top of this, sequence of sentences in a specific order makes a meaningful document and this document can be associated with a genre or a topic. Research in language modeling has reached substantial maturity with many pre-trained Word2Vec models (CBOW, Skip-gram) and pre-trained embeddings. In the domain of telemetry the sequence specific embeddings needs to be learnt organically and no-prior knowledge exists. We leverage learnings from 3-dimensional data modeling frameworks \cite{bert,attention} specifically in the context of positional encoding, localized and focused attention techniques and leverage them in our interpreter network. 
None of these contributions directly apply to us as we have dual training objectives and non symmetric data dimensions which warrants a new scrutiny. 
\section{CognitionNet Formulation}
\label{sec:cognitionNet}
CognitionNet is a collaborative framework comprising of two networks. The first network of CognitionNet is a sequence interpreter which can be assumed as process of mining the ``micro patterns''. Micro patterns  provide a high dimensional representation for continuous game play sequence. This process can be seen analogous to obtaining sentence level embeddings in NLP domain. The second network leverages the combined wisdom of the translated micro patterns per player to classify the overall pattern as a sustainer, burnout or a chrunout in this case. This is analogous to a topic discovery from sentence level embeddings for a document in NLP. 

\begin{figure*}
  \includegraphics[width=0.8\linewidth]{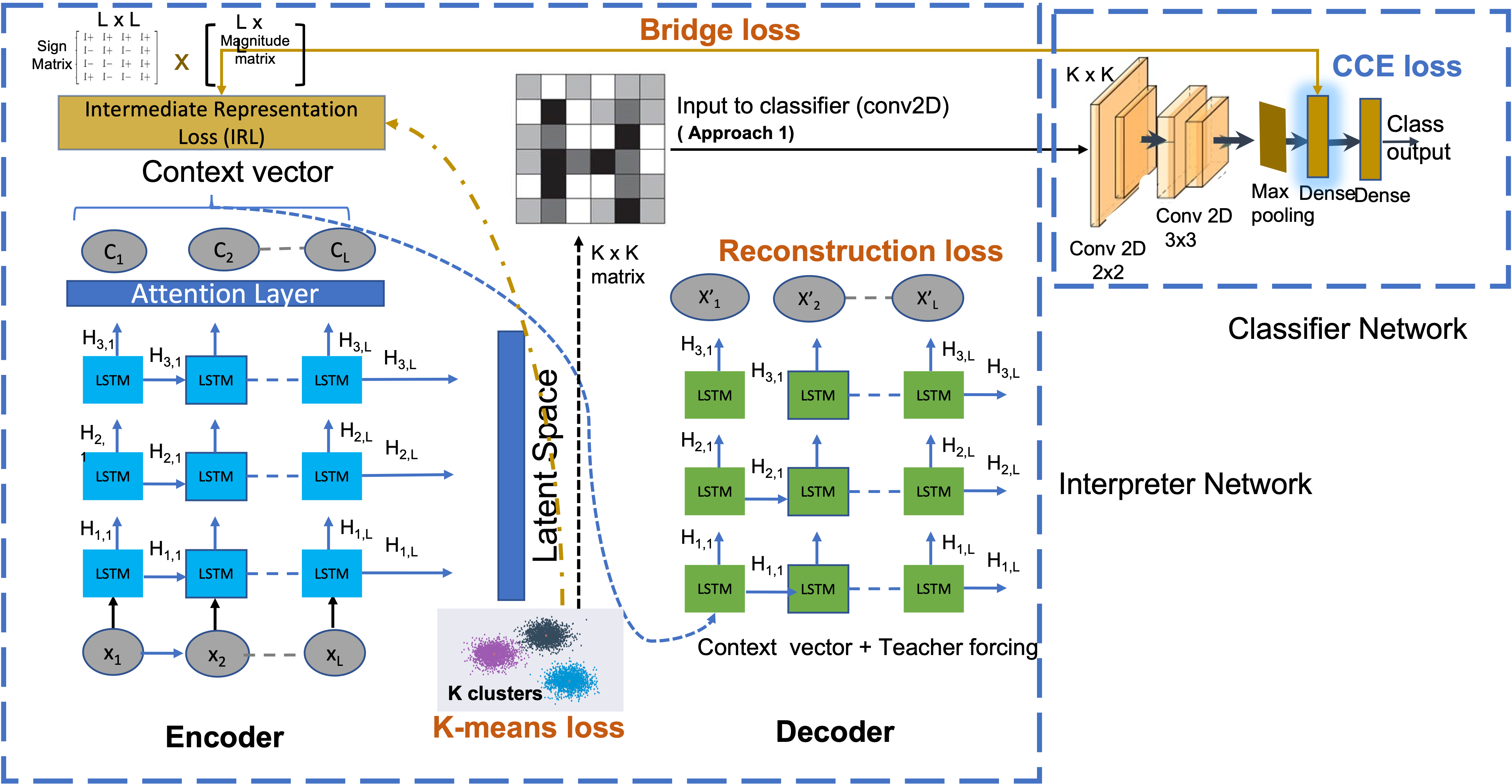}
  \caption{CognitionNet Architecture and Data/Loss Flow}
  \label{fig:cognitionNet}
\end{figure*}

CongnitionNet consists of a sequence to sequence (seq2seq) encoder-decoder based deep neural network as its first network ; and a supervised  convolutional neural network as its second network which is customizable to any other classifier network. The seq2seq network is being referred to as the \textbf{interpreter} and the convolution network is being referred to as the \textbf{classifier}. The general structure of CognitionNet is illustrated in Figure \ref{fig:cognitionNet}. The data flows through cognitionNet. Sequences are interpreted as micro patterns and micro patterns are further combined to predict player's class. Further in this process, micro patterns are used to derive explainable game behaviours. Progression in player's Game behaviors is \textit{aggregated} (e.g., transitions across patterns) to discover play styles. 

\subsection{Data Dimensions:}
\label{data_dimensions}
CognitionNet receives 3-dimensional input samples. Let \textit{N} be the total number of samples divided across \textit{C} labels or classes of players in our case. Table \ref{tab:data_dimension} illustrates the input and output dimensions of the interpreter and the classifier network. Each player's sample consists of an ordered sequence of sequences, order being important as it signifies its progression. For simplicity, let us assume we have total of \textit{S} sequences per player sample.  Each sequence is of length \textit{L}, L is the total number of games played at stretch by a player in a sequence.  Each game is represented with several features which could be numeric, Boolean or categorical. Let F be the total number of features per game and remains fixed throughout for all samples.  Table \ref{tab:game_features} lists a few of the game features. Each sequence may start at any time of the day as per player's active time on the platform. While the feature dimensions per game are fixed, the value of L varies from one sequence to another for a given player as well as across all players. While modeling, value of L is fixed across all sequences and classes. This limitation is addressed by configuring L to a 95\%le value of the distribution. Missing game entries are zeroed, analogous to handling shorter sentences in the NLP domain.  

\begin{table}[t]
  \centering
  \tiny
 \begin{tabular}{|c| l| l|}
 \hline
  Network   &  Input  &  Output \\
 \hline  
  Interpreter & (B1,~sequence length (L),~Features (F)) & (B1,~ Latent dimension (M)) \\
  \hline
  Classifier & transformed - (B2,~number of clusters (K),~K) & (B2,~ 1 (one of the classes)) \\
  \hline
\end{tabular}
 \caption{Data Dimensionality in CognitionNet: B1 and B2 being batch sizes of the two networks; B1 = S x B2 } 
 \label{tab:data_dimension}
\end{table}

\subsection{The Interpreter Network}
The interpreter network is quite identical to the one proposed in ``Deep Temporal Clustering Representations’’ (DTCR)  paper \cite{DTCR}. Our goal here is to perform unsupervised clustering on the sequences and obtain their cluster representations. The encoder is a 3-layered stacked LSTM network. It maps original time series into a latent space of representations. These representations are used to reconstruct the input data via decoder. Along with the representation loss, additionally, a K-means objective is integrated into the model to guide the representation learning. The core idea is to learn representations in an unsupervised manner via specific learning objectives. DTCR paper additionally proposes a fake-sample generation strategy and auxiliary classification task to enhance the ability of encoder. In the context of our working example, we do not have the concept of fake samples. Secondly, our proposed method is quite generic, the concept of fake samples is heavily dependent on the domain. 

The interpreter does not differentiate between sequences which are a part of the same sample vs. those belonging to two different samples. The input volume to the interpreter is N x S and shape of each input is L x F.  The output of the interpreter is a sequence cluster representation which is \textit{M} dimensional, which includes the hidden state outputs of the last time steps across all the stacked LSTM layers. This is in-tune with \cite{DTCR}. 
The decoder is also an identical LSTM network with 3 stacked layers. Along with the latent representation, each layer is being fed with context vector calculated using attention mechanism. 


\subsection{The Classifier Network}
\label{classifier_inputs}
The cluster representations of sequences in a sample are ordered in their original order prior to feeding them to the classifier. The shape of a data sample is (S,M) where S is number of sequences and M is the cluster representation's dimension for each sequence. Classifier learns via the supervised classification loss using the labels. In CognitionNet, additionally, the classifier is enabled to negotiate over cluster representations with the interpreter network. This is permissible as the cluster representations are 1) interpretations of the encoder and there is no absolute ground truth and 2) interpreter learns these interpretations only on k-means loss and the reconstruction loss which is not aligned with the classifier loss.  

\subsubsection{Input Mapping Function:} 

Interpreter generates the cluster representations for each sequence. We detach the output of the encoder from the input to the classifier. Our idea is to make choice of the classifier architecture agnostic to CognitionNet architecture and empower the user of the framework to incorporate domain specific wisdom to translate the progressive order of the  cluster representations for class prediction. For instance, in our context the cluster representations of sequences in a sequence can be classified using the following 3 different approaches.
\\
\noindent\textbf{Approach 1:} Transforming cluster representations to their corresponding cluster identifiers by performing a k-means operation on them as discussed in \cite{DTCR}. Here the value of K, is preconfigured while training the interpreter network. Further, only considering their consecutive transitions (sequence i to sequence i+1), wherein we model player's game behaviour transitions as a Markov process. In this transition or an adjacency matrix  value in each cell[i][j] denotes the number of times player moved from cluster i to cluster j in consecutive sequences. This normalized matrix is analogous to a ``heat map'' can be fed to a convolution based supervised classifier. The input dimension to the classifier would be K x K for each player's sample.\\
\noindent\textbf{Approach 2:} Directly passing sequential cluster representations to a LSTM/RNN based supervised classifier. The input dimension to the classifier would be S x M for each player's sample.\\
\noindent\textbf{Approach 3:} Obtain cluster identifiers for each sequence as in approach 1 and using frequency distributions of these sequences across k clusters as features. In this case the classifier could a simple ANN. The input dimension to the classifier would be K for each player's sample.

\subsubsection{Need for a Bridge Loss:}
In our use case we found approach 1 of combining the micro patterns quite appropriate keeping in mind the Markov assumption. However, this input mapping and using a convolution based classifier would make CognitionNet a very stringent framework. The concept of adjacency matrix may not apply in other use cases and so is the use of convolutions for classification. Secondly, the possibility of loss back-propagation would go away if the input is transformed before feeding to the classifier. This is because the logic of creating adjacency matrix would be outside of encoders loss computation graph. So we back-propagate classification errors from the classifier to the encoder. This brings in the need for bridge loss, which we will discuss shortly.

\subsection{Loss Functions:}
Given a set of N data samples; D = {$z_{1}$, $z_{2}$, ..., $z_{N}$}. Each sample $z_{i}$ contains S ordered sequences; X = ($x_{1}$, $x_{2}$, ..., $x_{S}$). Each $x_{i}$ is a sequence of L x F dimensions: ($x_{i,1}$, $x_{i,2}$, ...$x_{i,L}$) and $x_{i,1}$ = ($x_{i,1,1}$, $x_{i,1,2}$, ...., $x_{i,1,F}$). Each sequence is an input to the interpreter network. It defines non-linear mappings $f_{enc}$ : $x_{i,j}$ $\to$ $h_{i,j}$ and $f_{dec}$ : $h_{i,j}$ $\to$ $\hat{x}_{i,j}$ where each  . $f_{enc}$, $f_{dec}$, denotes the encoding and decoding process, respectively. Each $h_{i,j}$  $\in$ $\mathbf{R}^{M}$ is the M-dimensional latent representation of a sequence $x_{i,j}$, defined by:
$h_{i,j}= f_{enc}(x_{i,j})$. 
The after decoding we can obtain the output $\hat{x}_{i,j}$ , where $\hat{x}_{i,j}$ $\in$ $\mathbf{R}^{L}$ is given by: 
$\hat{x}_{i,j} = f_{dec}(h_{i,j})$. 
We use Mean Square Error (MSE) as the reconstruction loss, which is defined by:
\begin{equation} \label{reconstruction}
\mathcal{L}_{reconstruction}  = \frac{1}{L}\sum_{i=1}^{L} \lVert{x_{i,j} - \hat{x}_{i,j}}\rVert^2
\end{equation}
where j $\in$ (1,...,F). Although the learned representations by reconstruction loss capture the informative features of the original time series, they are not necessarily suitable for the clustering task as observed in \cite{DTCR}. To enable the learned representations to form cluster structures the network learning is guided through k-means.
Following the guidance from \cite{DTCR,spectral}, minimization of K-means is reformulated as a trace maximization problem associated with the Gram matrix $H^{T}$ $H$, which possesses optimal global solutions without local minima. Spectral relaxation converts the K-means objective into the following problem:
\begin{equation} 
\mathcal{L}_{K-means}  = Tr(H^{T} H) - Tr(F^{T} H^{T} HF)
\label{trace_loss}
\end{equation}
where Tr denotes the matrix trace. F $\in$ $R^{NxS,K}$ is the cluster indicator matrix. In reality, not all sequences of all samples are trained in one go but are split into multiple batches. The entire implementation is as per \cite{DTCR} and similar process of learning the H matrix and updating the F matrix on every $I^{th}$ iteration is being followed. We have experimentally arrived at this value of I as 10. More details are in the supplementary material.

\begin{table}[t]
  \centering
  \small
  \begin{tabular}{ | c | l | }
\hline
Dimension & Feature \\
\hline
\multirow {3}{*}{Pre-Game} & Type of Rummy Format to play  \\
& Entry fee paid while entering the table\\
& Deposited/ Withdrawn money into the wallet\\
\hline
\multirow {3}{*}{In-Game Choices} & Play slower than usual for focus \\
& Invalid declaration of the win \\
& Continue to play on bad cards (drop adherence)\\
\hline
\multirow {3}{*}{Outcome} & Won vs. lost the game \\
& Amount of money actually lost\\
\hline
\end{tabular}
  \caption{Sample Game Features: Drop adherence evaluates player's ability to judge if the initial rummy cards qualify as a bad hand \cite{kddgames,pakddgames}}
  \vspace{-0.3in}
  \label{tab:game_features}
\end{table}

\subsection{Bridge Loss:}
The Encoder outputs learnt cluster representations of all the S sequences for each sample. These inputs are transformed based on the approach 1 above, prior sending to the classifier.
This transformation happens outside the loss computation graph of the interpreter network. In order to facilitate the encoder to watch the classification errors, we derive intermediate representation loss (IRL) from the cluster representations of sequences. We ``\textbf{bridge}'' them with the state of the penultimate neural network layer of the classifier. This helps encoder to watch and understand the flow of errors. 
Bridge loss has two components: \textbf{1:} calculation of \textbf{I}ntermediate \textbf{R}epresentation \textbf{L}oss (IRL) and \textbf{2:} bridging it with classifier layer weights, called as Feature Matching (FM) step. 
\subsubsection{IRL formulation:}
 Each of the S sequences of a player $U$ are M dimensional. We represent all transitions of cluster representations between S sequences as matrix $H_{U}$ of shape (SxM, SxM) for a player U. We compute Cosine similarity across the player's sequences, to understand how these sequences are different or similar from each other. This magnitude matrix, (S,S), is represented as:
$MAG_{U}  = H^{T}_{U} H_{U}$. 
$H_{U}$ is a subset of the Gram matrix, previously discussed, since it is at a player and her sequences level. The second component is the interpretation of the sign of this magnitude as a penalty or a reward. This Penalty Reward sign matrix helps to bring out distinct behavioral clusters by giving penalty to distinct cluster transitions. Essence of this matrix is to create distinct transition paths for Sustainers, Burnouts and Churnouts, which in turn helps in creating localized patterns with varied intensities in the adjacency matrix. Let $SIGN_{U}$ be the sign matrix of shape (S,S) and defined as:
\begin{equation} \label{sign_matrix}
    SIGN_{U}[i][j] = 
\begin{cases}
    +1 (penalty),& \text{if } CL(H_{U}[i]) \neq CL(H_{U}[j]), \\
    & \hspace{2pt} \text{for i,j} \in \text{\{1,...S\}}~and~ i \neq j\\
    -1 (reward),              & \text{otherwise}
\end{cases}
\end{equation}
CL($H_{U}[i]$) is the cluster identifier of cluster representation for the $i^{th}$ sequence of a player U, via K-means. Negative sign indicates a reduction in loss and hence analogous to a reward. Transitions into the same cluster indicates move towards homogeneity and hence incentivized as a reward. Finally IRL of a sample is a directional matrix of shape (S , S) obtained as: 
$IRL_{U}  = MAG_{U} ~x ~SIGN_{U}$.
The $IRL_{U}$ matrix is reduced to (S,1) space by passing it through a trained Softmax activations for summarizing it to 1 value per sequence. 

\subsubsection{FM formulation:}
\begin{figure}
  \includegraphics[width=0.45\textwidth, height=4cm,keepaspectratio=true]{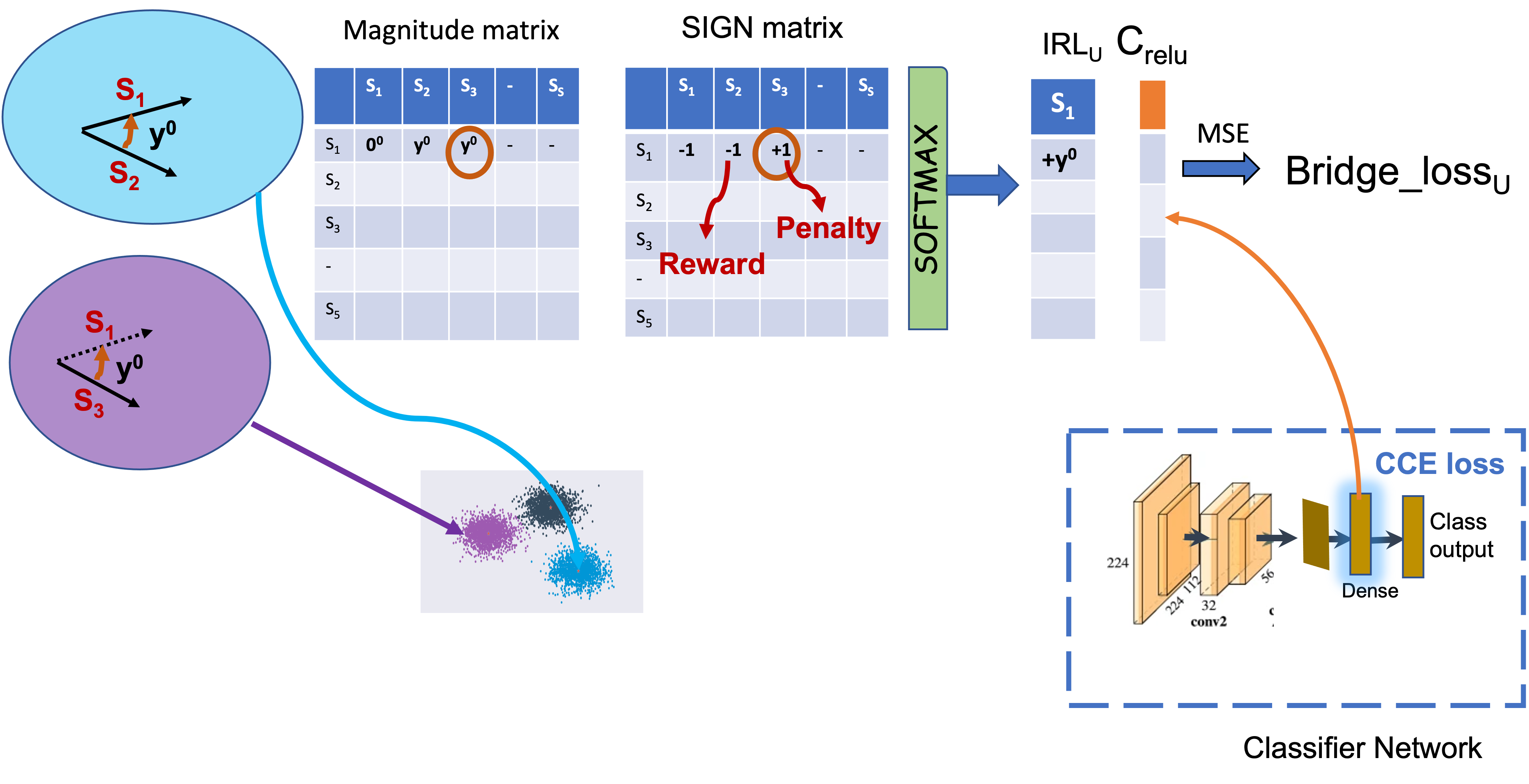}
  \caption{Bridge Loss Calculation Flow}
  \label{fig:FCM}
  \vspace{-0.2in}
\end{figure}

The IRL matrix for each player tries to match the discriminative features which helps in classification. This is incorporated in the network using the Mean Squared error between the IRL and penultimate layer ($\mathcal{C}_{Relu}$) of the classifier network. 
This technique of feature matching is inspired from the work on training a  generator in \textit{Open AI} \cite{fm}. We incorporate a similar set up where the interpreter network is given a new objective which prevents the classifier from over training over multiple collaborative epochs. At each collaborative epoch the interpreter is nudged (via bridge loss) to generate a latent space, which helps the classifier to understand and classify the data on varied latent spaces. Final bridge loss per sample is defined as:
\begin{equation} \label{bridge_loss}
 {Bridge\_loss}_{U} =  \frac{1}{S}\sum_{i=1}^{S} \lVert{\mathcal{C}_{Relu} - IRL_{U}}\rVert^2
\end{equation}
The training of the classifier is disabled in every alternative collaborative training step, analogous to training of a GAN framework \cite{gan}. Figure \ref{fig:FCM} depicts the various components and the steps in calculating the Bridge loss. Overall, in CognitionNet the interpreter loss is defined as:
\begin{equation} \label{loss_interpreter}
\begin{split}
\mathcal{L}_{Interpreter} = \beta ( \mathcal{L}_{reconstruction} & + \frac{\lambda}{2} ~    \mathcal{L}_{K-means} ) \\
& + \hspace{-1pt}( 1-\beta) {Bridge\_loss}
\end{split}
\end{equation}
Where the Bridge\_loss is averaged over the various players in a batch. Value of $\lambda$ was experimentally derived and was fixed at 0.5. The classifier loss depends on the type of the classifier in our case it is the categorical cross entropy loss (CCE). 
\section{Evaluation:}
In this section we present an evaluation of CognitionNet to predict the engagement class of the player. In the later section we will discuss the actionable explainations that evolve from CognitionNet which  deliver understanding of the various \textit{play styles} of these players. We use the same dataset as discussed in the Section \ref{background}. 

\subsection{Model Training:} 
In the first collaborative training epoch the interpreter network is trained with only the trace and the reconstruction loss on sequences for a batch of players. During the classifier training loop, adjacency matrix is created per sample as per approach 1  discussed in Section \ref{classifier_inputs}. The classifier is trained on its own supervised loss (CCE). In the next collaborative epoch on wards, while the interpreter is training, bridge loss is back propagated to the interpreter network. Now, the encoder starts adjusting its clusters to reduce the bridge loss along with the other two losses. The Classifier in the subsequent step under the same collaborative epoch sees improvement in its input and further trains itself. This collaborative alternate training of interpreter and classifier via bridge loss is continued for 5 - 10 collaborative epochs, until saturation is observed with respect to the bridge loss. We experimentally derive $\beta$ and fix it to 0.3. The number of collaborative epochs to train are dependent on the dataset.

\subsection{Results:}
\subsubsection{Classification Accuracy:}
 We performed experiments to validate the sanity of our network's architecture and the effectiveness of the bridge loss with the two variants of custom classifier and also against another custom solitary classification network. The two variants of the classifiers were - custom transition matrix (TM) based classifier (CognitionNet - TM) and the sequential (S) LSTM based classifier (CognitionNet - S). Both these classifiers were trained via bridge loss. The encoder's output is transformed appropriately to operate these classifiers.  The two classifier models have a competitive performance and have been primarily showcased to highlight the classifier agnostic loss propagation and training in CognitionNet.
 
\begin{table*}
  \centering
  \footnotesize
 \begin{tabular}{||c| c| c|c| c|c| c||}
 \hline
  Model &  Sustainer R mean\%  & Sustainer P mean\% &  Burnout R mean\%  & Burnout P mean\% &  Churnout R mean\%  & Churnout P mean\%\\
 \hline  
  CognitionNet - TM & \textbf{82.56} & \textbf{85.54} & \textbf{61.54}  & \textbf{84.21} & \textbf{94.32} & \textbf{86.01} \\ 
  CognitionNet - S & 36.05 & 52.54 & 23.08 & 21.05 & 72.16 & 64.14\\
  DTCR-alike + Conv  & 15.8 & 34.2 & 3.08 & 12.4 & 85.3 & 57.6  \\
  \hline
\end{tabular}
 \caption{Performance of CognitionNet with Variants and DTCR-alike: P - Precision, R - Recall} 
 \vspace{-0.2in}
 \label{tab:classfication_results}
\end{table*}
We claim that classification loss propagation via the bridge loss and the collaborative training helps to bring in separability between the three player classes. In another experiment we preserve the interpreter which is more like the DTCR network. We also attach an auxiliary classifier as in \cite{DTCR}. The fake samples are created in the same way from original sequences, for the sake of comparison. This DTCR-alike network, encodes all the sequences per player. Now on the transformed input we apply classification by disconnecting the bridge loss. So essentially, the two networks are no longer connected. We observe that the performance of the  network, DTCR-alike + Conv  drastically reduces for all three classes  in comparison with CognitionNet - TM, where the classifiers are the same. We see similar results for other classifiers as well. 

\begin{figure}
    \subcaptionbox{CCE loss}{\includegraphics[width=0.15\textwidth]{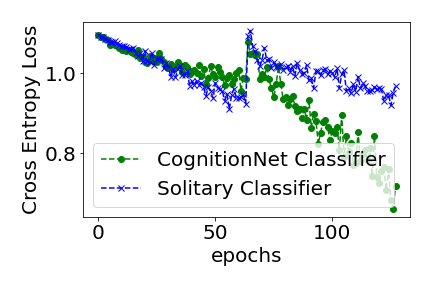}} 
    \subcaptionbox{Trace loss}{\includegraphics[width=0.15\textwidth]{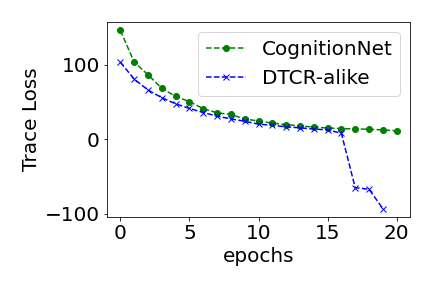}} 
    \subcaptionbox{Reconstruction loss}{\includegraphics[width=0.15\textwidth]{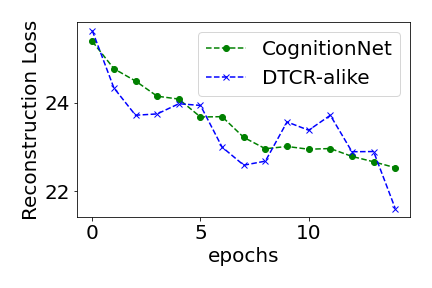}}
    \caption{Comparison of Loss Trajectories in CognitionNet vs. DTCR-alike + Conv }
    \label{fig:losses}
    \vspace{-0.2in}
\end{figure}

The classifier in cognitionNet collaborates with the interpreter and after the first collaborative epoch which is essentially after the first 60 epochs of the classifier, the losses start further dropping as shown in the Figure \ref{fig:losses}(a). The drop in the first 60 epochs of the classifier is solely due to its own CCE loss. The steep drop from the epoch 61 onwards is attributed to the bridge loss. To avoid any bias the solitary classifier is also trained twice for 60 epochs each time. We see that loss trajectory simply repeats. This is because after fully training the DTCR-alike, the encoding space doesn't shift further. The solitary classifier follows the same CCE loss trajectory once again.
The trace loss in DTCR-alike drops well and steadily across the training epochs. Note: trace loss can be negative as per the formulation in \cite{DTCR}. In case of CognitionNet the drop is hindered from the second collaborative epoch and it settles at a higher value (Figure \ref{fig:losses}(b)). The identical case is seen in the reconstruction loss where reconstruction accuracy also slightly suffers due to bridge loss compared to that in DTCR-alike (Figure \ref{fig:losses}(c)). The epochs in the figures are the respective network's local training epochs. 


\subsubsection{Class Transition Homogeneity - For Explainability:}
\begin{figure}
  \includegraphics[width=\linewidth]{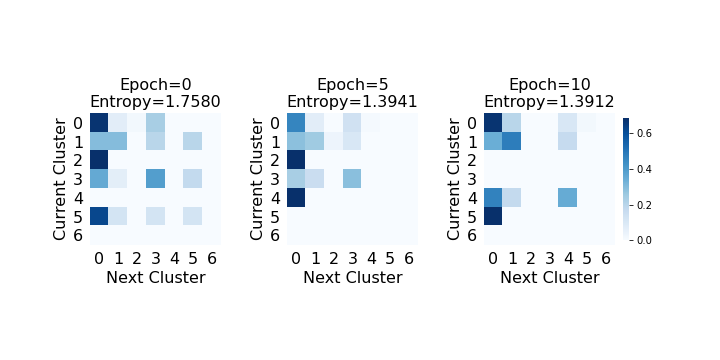}
  \vspace{-0.5in}
  \caption{Evolving Adjacency Matrix with dropping Entropy}
  \label{fig:heatmap_cognitionnet}
  \vspace{-0.2in}
\end{figure}

\begin{figure}
    \subcaptionbox{prior to bridge loss}{\includegraphics[width=0.15\textwidth]{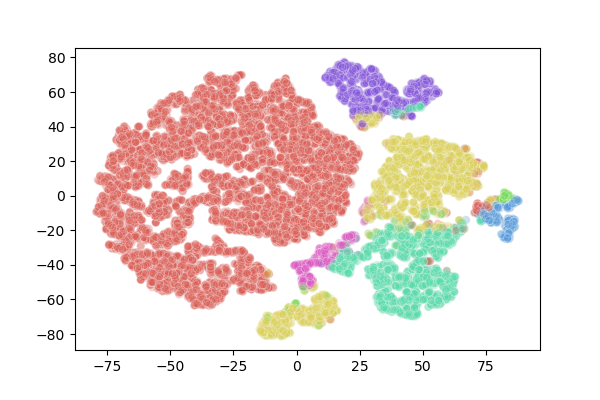}} 
    \subcaptionbox{intermittent epoch}{\includegraphics[width=0.15\textwidth]{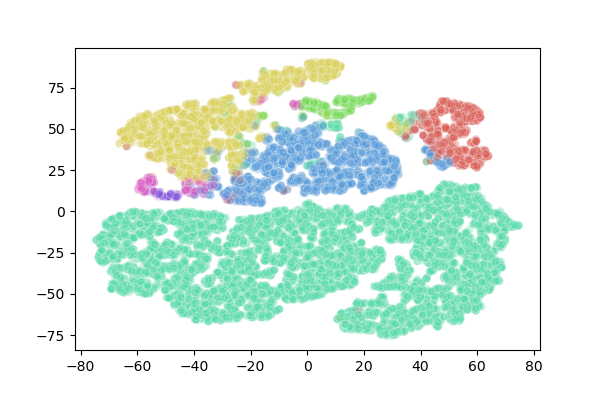}}
    \subcaptionbox{final epoch}{\includegraphics[width=0.15\textwidth]{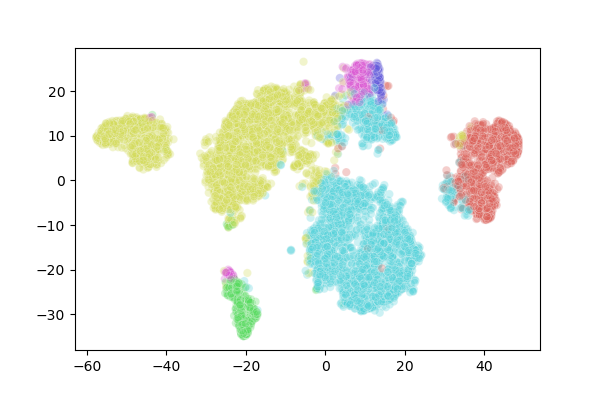}}
    \caption{Cluster formation before and after bridge loss }
    \label{fig:TSNE}
    \vspace{-0.2in}
\end{figure}

We earlier discussed the objective function of the bridge loss is to bring in homogeneity of transitions across all player samples for a class. One of the ways to quantify this property is to calculate \textit{entropy}. High entropy in any adjacency matrix pertaining to a player implies more uncertainty of transitions, less predictability and implies its sequences are distributed across multiple clusters . Figure \ref{fig:heatmap_cognitionnet} shows how the adjacency matrix evolves slowly for a player in CognitionNet. The initial high entropy is seen at lower trace loss. As the negotiations over bridge loss kick in the entropy starts reducing leading to lower number of cross cluster transitions. In the figure the epochs refer to the collaborative training epochs.
Taking the same point a little further Figure \ref{fig:TSNE} shows TSNE plots of cluster formations. Figure \ref{fig:TSNE}(a) shows 7 cluster boundaries upon full training of the interpreter (for k = 7). The bridge loss is not introduced yet. These boundaries are optimized on trace loss and appear more spherical then the latter ones. In \ref{fig:TSNE}(b), the interpreter is at the epoch post introduction of the bridge loss. This is evident in the irregular and nearly overlapping cluster boundaries. As the collaborative training proceeds there is movement in-out of the clusters leading to irregular shaped clusters. In the final, fully trained cluster representation shown in Figure \ref{fig:TSNE}(c) we omit the dense cluster and highlight the converged shapes of other 6 clusters. These irregularities are desirable for achieving homogeneity, as also evident through transition entropies in Figure \ref{fig:heatmap_cognitionnet}. Note: one shortcoming with using K-means is that the cluster identifiers constantly change across epochs.  The cluster identities are fixed only after the complete training. For these reasons the matrices in Figure \ref{fig:heatmap_cognitionnet} show inconsistent movement of densities across cells.
\footnote{As of today, there are no other open datasets with sequence of sequences to validate this further and hence we publish our source code as well at the data \cite{github}.}


\section{Game Behaviour and Play styles}

CognitionNet's trained model maps each sequence to a cluster. We analyze the raw sequence data per cluster to delve into the aspects of game behaviour and play styles via answering certain questions: 1) What does each cluster signify? 2) How do these clusters map into the various player classes? and 3) How do we discover play styles and associate them with the engagement classes? In the context of our use case the questions can be reformulated as: 1) Game Behaviour Identification; 2) Behaviour Transition Understanding and 3) Play style Discovery.

\subsection{Dominant Game Behaviour Identification:}
Prior to building CognitionNet, we made certain hypotheses by deploying logical reasoning and drawing inference from established theories \cite{playpersona,bighex} and observed user behaviour on the platform. We hypothesize certain attributes for the cash game players:\\
\noindent\textbf{Utilitarian Outlook (UO):} Players with this attitude are well aware of their game skills - capabilities and limitations. They carefully select games and play those games where they are confident of winning. They are practical, know when to stop and don't overdo.\\
\noindent\textbf{Commitment and Focus (CF):} They go step by step while selecting and addressing challenges. They show high focus and commitment to achieve desired results.\\
\noindent\textbf{Passive Aggression (PA):} These strongly believe that their true worth and potential has not yet been acknowledged. Believe in risk taking for deriving gratification.\\
\noindent\textbf{Gamblers Delight (GD):} These are by and large high risk takers, moving from optimism to over optimism. Take chances to maximize returns and stay adamant with the choices. \\
\noindent\textbf{Conservative (CS):} They exhibit a trait of self-doubt and always stay within their own comfort zone when it comes to choosing entry fees, game table sizes. They avoid playing games which ``they perceive'' to have lower (but actually above median) chances of winning and hence sometimes come across as a lacking trust upon their own skills.

\begin{figure}
   \includegraphics[width=0.15\textwidth]{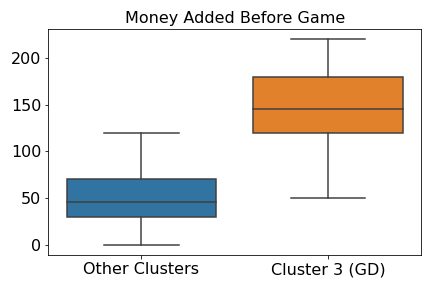}
    \includegraphics[width=0.15\textwidth]{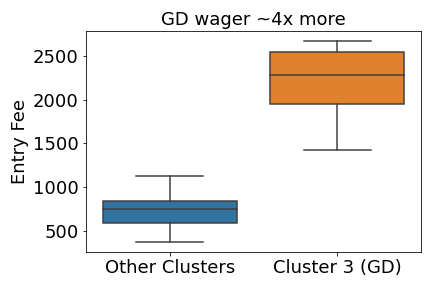}
    \includegraphics[width=0.15\textwidth]{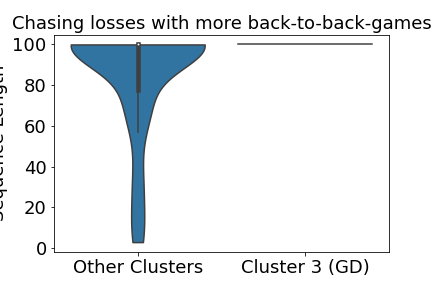}
    \medskip
    \includegraphics[width=0.15\textwidth]{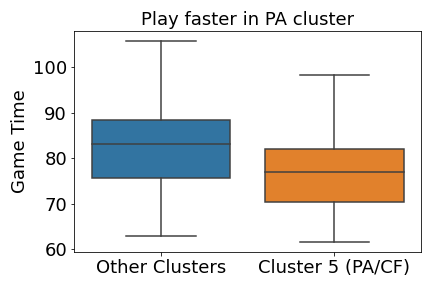} 
    \includegraphics[width=0.15\textwidth]{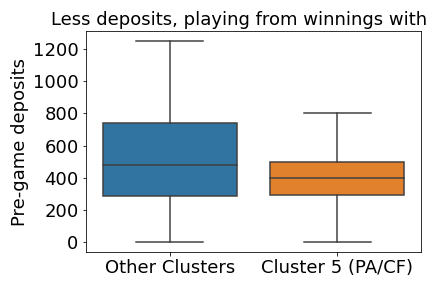}
    \includegraphics[width=0.15\textwidth]{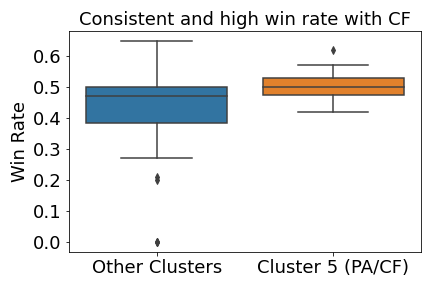}
    
    \caption{Box plots showing differences in features for GD and PA with CF clusters }
    \label{fig:boxplots}
    \vspace{-0.2in}
\end{figure}

We focus on top 4 densely populated clusters out of the total 7 that the network found and their corresponding dominant game behaviours.\\
\noindent\textbf{Cluster 0 $\Rightarrow$ \textit{UO}:} Sequences have smaller number of continuous games (< 40\% ) ; indicate longer (> 20\% ) time gaps from previous game finish to next game start - analogous to taking breaks, games are played with much lower entry fee (< 30-50\% ) and as an effect sequences indicate relatively lower net losses across all games in it (< 10-30\%). The figures of comparison here are against the average values of the same metrics in other clusters. This goes very well with our earlier definition of UO, where they are confident but careful and know when to stop.  \\
\noindent\textbf{Cluster 4 $\Rightarrow$ \textit{CS}:} Sequences have high number of continuous games (>100 games) but the choice of entry fee, game format remains monotonically the same, showing little or no openness to new experiences. Players take no breaks between game plays and the focus is on winning and withdraw money immediately after winning. \\
\noindent\textbf{Cluster 5 $\Rightarrow$ \textit{PA with CF}:} Sequences show tendency to play at wide range of entry fees, but with constant and high drop adherance indicating high focus at all times. As discussed in \cite{pakddgames}, high drop adherance indicates good ability to judge the quality of rummy hand when cards are dealt. Entry fees are raised progressively upon winning (> 40\% of the times) and kept unchanged after a game loss. Tendency to play at high entry fees also incurs negative net gain or loss. But yet, player does not make fresh wallet deposits but plays from winnings showing growth with commitment. Move to larger tables (rummy with more players on the table) is only seen on wins.\\
\noindent\textbf{Cluster 6 $\Rightarrow$ \textit{PA moving towards GD}:} Sequences reveal games on multiple tables at the same time. This cluster contains the sequences with rummy \textit{rumble} format being played. In rumble the player subscribes for beating a target score while playing rummy to win cash award and appear on the performer's list (can be seen as a path to social acceptance and recognition). While chasing targets players play with higher entry fees, for exceedingly large times with no breaks between games. More often money deposits to refill wallets are observed. Win rates on high entry fees are low and drop adherence declines subsequently on high losses indicating desperation and loss of judgement with a gambling high. There is, otherwise, no correlation found between playing rumble and leading into PA or GD attitudes. 
Figure \ref{fig:boxplots} shows some interesting statistics about Cluster 6 (GD) and Cluster 5 (PA with CF). Due to space limitations we had to omit depictions of many such observations.

\subsection{Behaviour Transition Understanding:} Interestingly, as hypothesized we see that players from a given prediction category mostly show stationarity and in-out transitions from only a limited set of clusters.  Study \cite{cog-playstyles} mentions that clustering and summarizing  behavioral time series does not mean to flatten the individual differences between players.  We too observe that each player delivers a slightly different story but it was mostly consistent with the overall theme of the class. For the purpose of demonstration in this paper, we picked a player per class to deliver a clearer, more organized representation of player's persona representing that class. 

\begin{figure}
    \centering
    \subcaptionbox{Player-S}{\includegraphics[width=0.15\textwidth]{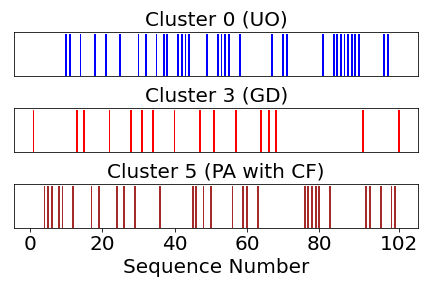}} 
    \subcaptionbox{Player-B}{\includegraphics[width=0.15\textwidth]{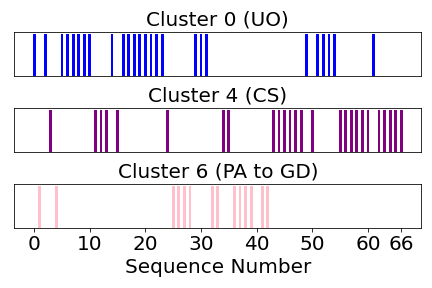}} 
    \subcaptionbox{Player-C}{\includegraphics[width=0.15\textwidth]{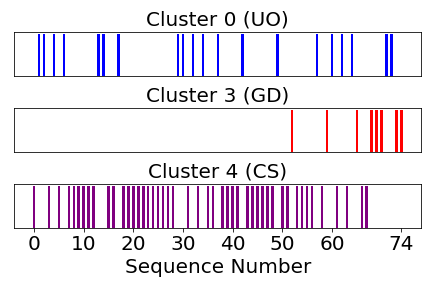}}
    \caption{Temporal Transitions Across Dominant Clusters }
    \label{fig:temporal_3_players}
    \vspace{-0.1in}
\end{figure}

\begin{figure}
    \subcaptionbox{Player-S}{\includegraphics[width=0.15\textwidth]{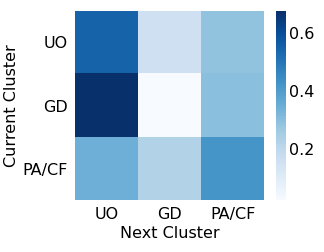}} 
    \subcaptionbox{Player-B}{\includegraphics[width=0.15\textwidth]{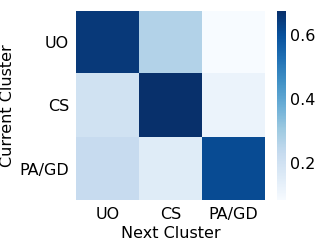}} 
    \subcaptionbox{Player-C}{\includegraphics[width=0.15\textwidth]{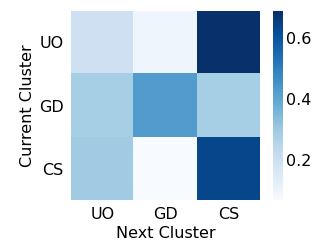}}
    \caption{Heatmap of Spatial Transitions Across Dominant Clusters }
    \label{fig:heatmap_3_players}
    \vspace{-0.1in}
\end{figure}

Player-S, Player-B and Player-C represent sequences of players from Sustainer, Burnout and a Churnout class respectively.
Figure \ref{fig:heatmap_3_players} shows transitions of the three players across these clusters during the entire observation period. Figure \ref{fig:temporal_3_players}  shows how over the observed time lines, the sequences of the players move between these clusters. While the first set of graphs show static behaviour, the second ones show temporal shifts. We summarize our observations as follows:
\noindent\textbf{Player-S}: Figure \ref{fig:heatmap_3_players} shows that this player has equal stationarity in the UO and the PA with CF clusters. 
\begin{wrapfigure}{l}{0.5\linewidth}
\includegraphics[width=\linewidth]{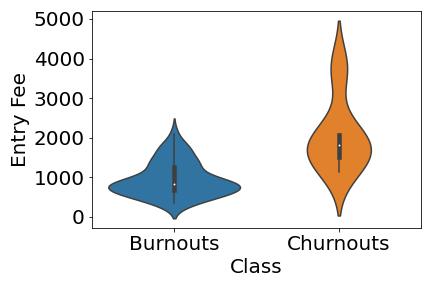}
\caption{Higher stakes of entry fees by overall churnouts in Cluster 4 $\Rightarrow$ CS}
\label{fig:violin_B_C}
\end{wrapfigure}
The transitions from UO to UO or to PA with CF clusters is equally likely but once in PA with CF cluster, the player tends to stick in there.  Figure \ref{fig:temporal_3_players} shows that temporally, in the initial periods the player had frequent PA and CF behaviours but over time, as she grew confident on the skills the utilitarian attitude kicked in. The cluster 3 is high GD, which alludes to the fact that this player does at times takes up to gambling but returns back to the regular CF attitude. 

\noindent\textbf{Player-B}: Similar corresponding graphs for a Burnout show that she is UO but at times gets conservative with choices, indicating self doubt and lack of desire to enjoy the playing and shifts the focus on winning (need to feel accomplished). This desire to win perhaps makes the player subscribe to rumble. At this point, Cluster 6 which is PA leading to GD kicks in, leading to further deterioration and probable burnout.

\noindent\textbf{Player-C}: Unlike for other two classes, for churnouts, we see mostly the after-effects rather than slow progressions of the causes. 
The particular churnout just before moving out of the platform shows heavy GD attitudes. She is mostly seen to be CS as such with monotonous tone of play, seeking for wins. UO attitude is less prominent and equally distributed throughout the observation interval. We validated that, this UO attitude is triggered by lower wallet balance and poor fresh deposits to continue playing.  It perhaps alludes to the fact that CS attitude might indicate waning interest of the player. Note: though both Player-B and Player-C fall into the CS cluster at some point in time, we indeed see that though both opt-in for monotonous and rigid choices there is clear separation between actual feature values where Player-C plays at ``higher'' stakes indicating aggression and subsequent losses leading to churn. Figure \ref{fig:violin_B_C} shows  overall distributions of entry fees in CS cluster for all churnouts vs. burnouts.

\vspace{-0.1in}
\subsection{Playstyle Discovery:}
\noindent\textbf{Sustainers as ``\textit{Masterminds}}'': This player can be seen to demonstrate good balance of UO which is important to prevent over-indulgence and equal appeal to PA keeping lazor focus on strategy and execution. This player is a \textit{master} player and rightfully sustains. 

\noindent\textbf{Burnouts as ``\textit{Trailing Daredevils}}'': This playstyle is all about the repeated urge to take risks where the stakes are high. This style excites them to take a plunge but they fail to withdraw (stickiness into CS). They generally play on the edge leading to frequent falls and feel the burn. 

\noindent\textbf{Chrunouts as ``\textit{Strugglers}}'': This indicates lack of strategy. Players are slowed down by the outcome of their own choices. They exhibit GD for ``one last try''. More CF with higher UO can protect them.


\section{Conclusion}
We introduce CognitionNet which is an assembly of two neural networks, where, the first network is focused on interpreting the first two dimensions into “micro patterns”. The second network has to leverage the collective wisdom of these micro patterns (i.e., third dimension) for a player to mine her play style, while grounding that towards predicting the engagement. These networks collaborate to refine the representations of the micro patterns and create robust definitions of play styles. However, since the input dimensions differ, we propose a novel formulation for loss back-propagation, referred to as the Bridge loss for the necessary collaboration. The dual objective allows CognitionNet to reveal several player psychology inspired decision making and tactics. 

\noindent \textit{CognitionNet Adoption:}  
The ability to automatically analyze sequence of sequences for a sparse and high dimensional game play data and predict playing styles per player would have tremendous benefits as summarized here:
\noindent\textbf{Prevent Player Fatigue:} 
Immediate game behavior identification can be used to nudge down an aggressive player with personalized recommendations like: playing with lower wage values, and suggestions on taking breaks.
\noindent\textbf{Personalized Targets:} We have seen that players nearing a burnout tend to subscribe to challenges and handle them with lower focus. 
Players with high probability of churn or a burn could be given lesser aggressive targets in games like rummy rumble.
\noindent\textbf{Identify a Rummy Buddy:} Players high on Focus and Commitment, specifically the \textit{masterminds}, can be offered to volunteer on the platform to raise awareness on healthy playing habits and provide simple tips and tricks of the game of rummy for the novice.
These initiatives are underway and would soon help in building an even more safe and competitive gaming experience.

\bibliographystyle{plain}


\begin{appendices}

\begin{figure*}
  \includegraphics[width=0.8\linewidth]{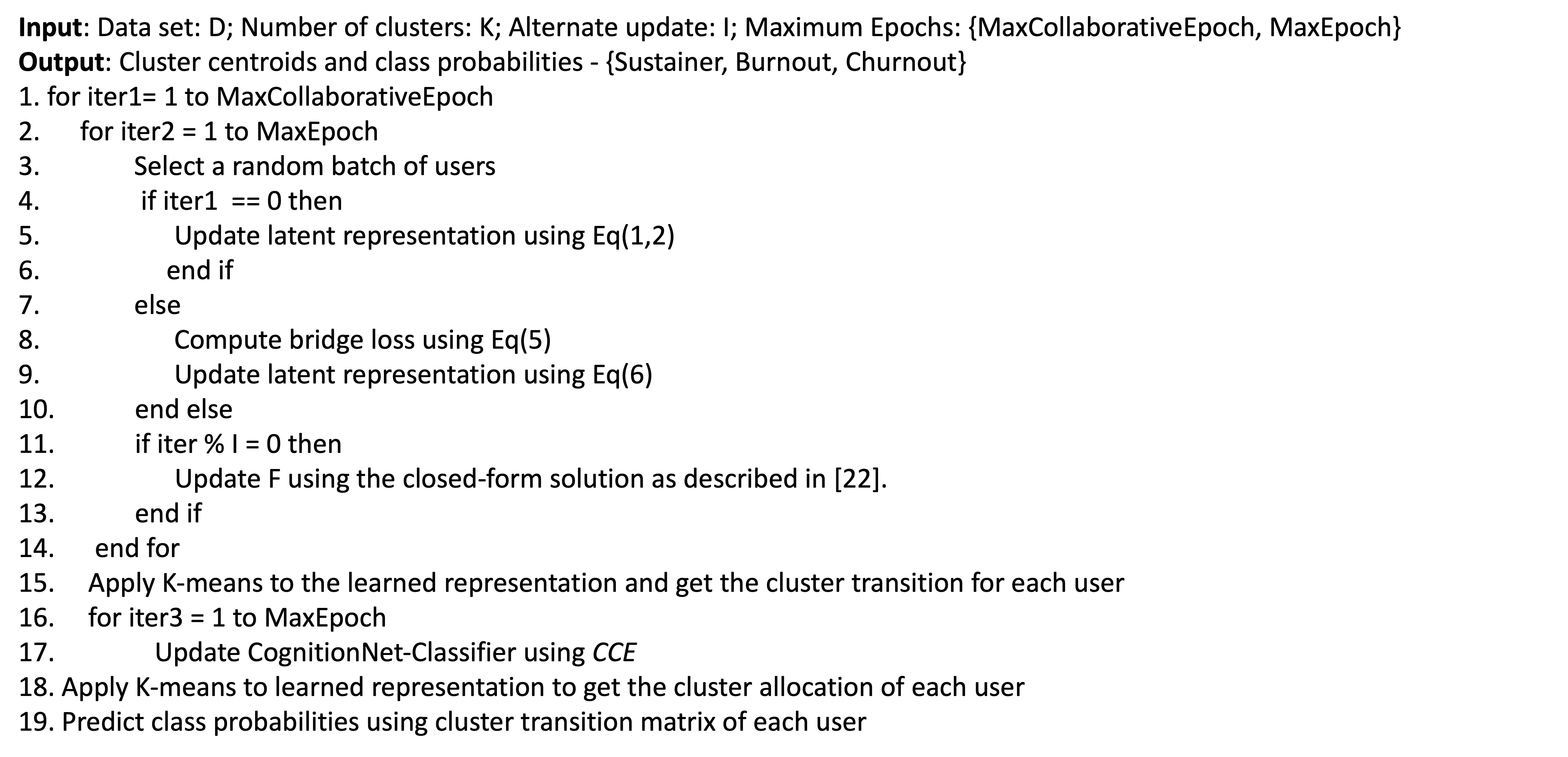}
  \caption{CognitionNet Algorithm/Pseudocode}
  \label{fig:pseudocode}
\end{figure*}

\begin{table*}
  \centering
  \footnotesize
  \caption{Variation in Performance with varying number of clusters(K) (mean - 5 runs) - (P - Precision, R- Recall)} 
 \begin{tabular}{||c|c|c|c|c|c|c||}
 \hline
  Parameter Value &  Sustainer R mean\%  & Sustainer P mean\% &  Burnout R mean\%  & Burnout P mean\% &  Churnout R mean\%  & Churnout P mean\%\\
 \hline  
  K = 4 & 69.77 & 60.00 & 55.77 & 39.19 & 60.23 & 75.71 \\\hline 
  K = 5 & 50.00 & 61.43 & 44.23 & 30.67 & 67.05 & 69.28 \\\hline 
  K = 6 & 53.49 & 64.79 & 40.38 & 30.43 & 69.32 & 70.11 \\\hline 
  \textbf{K = 7} & 82.56 & 85.54 & 61.54 & 84.21 & 94.32 & 86.01 \\\hline 
  K = 8 & 54.65 & 46.53 & 26.92 & 16.67 & 56.82 & 77.52 \\\hline 
  \hline
\end{tabular}
\end{table*}
\vspace{0.3cm}
\begin{table*}
  \centering
  \footnotesize
   \caption{Variation in Performance with varying $\lambda$ (mean - 5 runs) - (P - Precision, R- Recall)} 
 \begin{tabular}{||c|c|c|c|c|c|c||}
 \hline
  Parameter Value &  Sustainer R mean\%  & Sustainer P mean\% &  Burnout R mean\%  & Burnout P mean\% &  Churnout R mean\%  & Churnout P mean\%\\
 \hline  
  \textbf{$\lambda$} = 0.5 & 82.56 & 85.54 & 61.54 & 84.21 & 94.32 & 86.01 \\\hline 
  $\lambda$ = 1.0 & 65.12 & 62.92 & 50.00 & 36.62 & 67.05 & 76.62 \\\hline 
  \hline
\end{tabular}
\end{table*}
\vspace{0.3cm}

\begin{table*}
  \centering
  \footnotesize
  \caption{Variation in Performance with varying I (mean - 5 runs) - (P - Precision, R- Recall)} 
 \begin{tabular}{||c|c|c|c|c|c|c||}
 \hline
  Parameter Value &  Sustainer R mean\%  & Sustainer P mean\% &  Burnout R mean\%  & Burnout P mean\% &  Churnout R mean\%  & Churnout P mean\%\\
 \hline  
  I = 5  & 55.81 & 52.17 & 25.00 & 46.43 & 78.98 & 71.65 \\\hline 
  \textbf{I = 10} & 82.56 & 85.54 & 61.54 & 84.21 & 94.32 & 86.01 \\\hline 
  I = 15 & 65.12 & 57.14 & 28.85 & 42.86 & 75.57 & 73.48 \\\hline 
  \hline
\end{tabular}
\end{table*}
\vspace{0.3cm}

\begin{table*}
  \centering
  \footnotesize
  \caption{Variation in Performance with varying $\beta$ (mean - 5 runs) - (P - Precision, R- Recall)}  \begin{tabular}{||c|c|c|c|c|c|c||}
 \hline
  Parameter Value &  Sustainer R mean\%  & Sustainer P mean\% &  Burnout R mean\%  & Burnout P mean\% &  Churnout R mean\%  & Churnout P mean\%\\
 \hline  
  $\beta$ = 0.2 & 75.58 & 68.42 &  0.00 &   0.0 & 90.34 & 72.60 \\\hline 
  \textbf{$\beta$ = 0.3} & 82.56 & 85.54 & 61.54 & 84.21 & 94.32 & 86.01 \\\hline 
  $\beta$ = 0.4 & 66.28 & 72.15 & 43.75 & 53.85 & 74.43 & 76.61 \\\hline 
  $\beta$ = 0.5 & 55.81 & 63.16 & 51.92 & 42.86 & 73.30 & 73.71 \\\hline 
  $\beta$ = 0.6 & 51.16 & 68.75 & 63.46 & 36.26 & 65.91 & 72.96 \\\hline 
  $\beta$ = 0.7 & 61.63 & 60.92 & 55.77 & 34.52 & 59.66 & 73.43 \\\hline 
  $\beta$ = 0.8 & 63.95 & 63.22 & 51.92 & 27.84 & 52.27 & 70.77 \\\hline 
  $\beta$ = 0.9 & 65.12 & 69.14 & 34.29 & 69.23 & 52.27 & 71.88 \\\hline 
  \hline
\end{tabular}
\end{table*}
\vspace{0.3cm}

\end{appendices}

\end{document}